\documentclass[11pt,a4paper]{article}
\usepackage[hyperref]{acl2020}
\usepackage{times}
\usepackage{latexsym}

\usepackage{makecell}

\usepackage{times}
\usepackage{epsfig}
\usepackage{graphicx}
\usepackage{amsmath}
\usepackage{amssymb}
\usepackage{longtable}

\makeatletter
\def\thickhline{%
  \noalign{\ifnum0=`}\fi\hrule \@height \thickarrayrulewidth \futurelet
   \reserved@a\@xthickhline}
\def\@xthickhline{\ifx\reserved@a\thickhline
               \vskip\doublerulesep
               \vskip-\thickarrayrulewidth
             \fi
      \ifnum0=`{\fi}}
\makeatother

\newlength{\thickarrayrulewidth}
\usepackage{multirow}
\usepackage{booktabs}

\usepackage{pifont}
\newcommand{\cmark}{\ding{51}}%
\newcommand{\xmark}{\ding{55}}%

\newcommand{\xdownarrow}[1]{%
  {\left\downarrow\vbox to #1{}\right.\kern-\nulldelimiterspace}
}

\usepackage{xcolor}

\usepackage{colortbl}
\definecolor{Gray}{gray}{0.85}

\usepackage{microtype}

\aclfinalcopy 

\title{MART: Memory-Augmented Recurrent Transformer \\ for Coherent Video Paragraph Captioning}

\author{Jie Lei$^1$\thanks{\quad Work done while Jie Lei was an intern and Yelong Shen was an employee at Tencent AI Lab.} \quad Liwei Wang$^2$ \quad Yelong Shen$^3$\footnotemark[1] \quad Dong Yu$^2$ \quad Tamara L. Berg$^1$ \quad Mohit Bansal$^1$ \\ 
  $^1$UNC Chapel Hill \quad $^2$Tencent AI Lab Seattle USA \quad $^3$Microsoft Dynamics 365 AI \\ 
  \texttt{\{jielei, tlberg, mbansal\}@cs.unc.edu} \\
  \texttt{\{liweiwang, dyu\}@tencent.com}, \texttt{\{yeshe\}@microsoft.com}
  }

\begin{document}
\maketitle
\begin{abstract}
Generating multi-sentence descriptions for videos is one of the most challenging captioning tasks due to its high requirements for not only visual relevance but also discourse-based coherence across the sentences in the paragraph. 
Towards this goal, we propose a new approach called Memory-Augmented Recurrent Transformer (MART), which uses a memory module to augment the transformer architecture. 
The memory module generates a highly summarized memory state from the video segments and the sentence history so as to help better prediction of the next sentence (w.r.t. coreference and repetition aspects), thus encouraging coherent paragraph generation. 
Extensive experiments, human evaluations, and qualitative analyses on two popular datasets ActivityNet Captions and YouCookII show that MART generates more coherent and less repetitive paragraph captions than baseline methods, while maintaining relevance to the input video events.\footnote{All code is available open-source at \url{https://github.com/jayleicn/recurrent-transformer}}
\end{abstract}

\section{Introduction}\label{introuction}
In video captioning, the task is to generate a natural language description capturing the content of a video. Recently, dense video captioning~\cite{krishna2017dense} has emerged as an important task in this field, where systems first generate a list of temporal event segments from a video, then decode a coherent paragraph (multi-sentence) description from the generated segments. \citet{park2019adversarial} simplifies this task as generating a coherent paragraph from a provided list of segments, removing the requirements for generating the event segments, and focusing on decoding better paragraph captions from the segments.
As noted by \citet{Xiong2018MoveFA,park2019adversarial}, generating paragraph descriptions for videos can be very challenging due to the difficulties of having relevant, less redundant, as well as coherent generated sentences. 

Towards this goal, \citet{Xiong2018MoveFA} proposed a variant of the LSTM network~\cite{hochreiter1997long} that generates a new sentence conditioned on previously generated sentences by passing the LSTM hidden states throughout the entire decoding process. 
\citet{park2019adversarial} further augmented the above LSTM caption generator with a set of three discriminators that score generated sentences based on defined metrics, i.e., relevance, linguistic diversity, and inter-sentence coherence. Though different, both these methods use LSTMs as the language decoder. 

Recently, transformers~\cite{vaswani2017attention} have proven to be more effective than RNNs (e.g., LSTM~\cite{hochreiter1997long}, GRU~\cite{chung2014empirical}, etc.), demonstrating superior performance in many sequential modeling tasks~\cite{vaswani2017attention, zhou2018end, devlin2018bert, dai2019transformer, yang2019xlnet}. 
\citet{zhou2018end} first introduced the transformer model to the video paragraph captioning task, with a transformer captioning module decoding natural language sentences from encoded video segment representations. 
This transformer captioning model is essentially the same as the original transformer~\cite{vaswani2017attention} for machine translation, except that it takes a video representation rather than a source sentence representation as its encoder input.
However, in such design, each video segment caption is decoded individually without knowing the context (i.e., previous video segments and the captions that have already been generated), thus often leading to inconsistent and redundant sentences w.r.t. previously generated sentences (see Figure~\ref{fig:caption_example} for examples).
\citet{dai2019transformer} recognize this problem as \textit{context fragmentation} in the task of language modeling, where the transformers are operating on separated fixed-length segments, without any information flow across segments.
Therefore, to generate more coherent video paragraphs, it is imperative to build a model that can span over multiple video segments and capture longer range dependencies.

Hence, in this work, we propose the Memory-Augmented Recurrent Transformer (MART) model (see Section~\ref{sec:methods} for details), a transformer-based model that uses a shared encoder-decoder architecture augmented with an external memory module to enable the modeling of the previous history of video segments and  sentences. 
Compared to the vanilla transformer video paragraph captioning model~\cite{zhou2018end}, our first architecture change is the unified encoder-decoder design, i.e., the encoder and decoder in MART use shared transformer layers rather than separated as in~\citet{zhou2018end, vaswani2017attention}.
This unified encoder-decoder design is inspired by recent transformer language models~\cite{devlin2018bert, dai2019transformer, sun2019videobert} to prevent overfitting and reduce memory usage. 
Additionally, the memory module works as a memory updater that updates its memory state using both the current inputs and previous memory state.
The memory state can be interpreted as a container of the highly summarized video segments and caption history information.
At the encoding stage, the current video segment representation is enhanced with the memory state from the previous step using cross-attention~\cite{vaswani2017attention}. Hence, when generating a new sentence, MART is aware of the previous contextual information and can generate paragraph captions with higher coherence and lower repetition.

Transformer-XL~\cite{dai2019transformer} is a recently proposed transformer language model that also uses recurrence, and is able to resolve context fragmentation for language modeling~\cite{dai2019transformer}. 
Different from MART that uses a highly-summarized memory to remember history information, Transformer-XL directly uses hidden states from previous segments. 
We modify the Transformer-XL framework for video paragraph captioning and present it as an additional comparison. 
We benchmark MART on two standard datasets: ActivityNet Captions~\cite{krishna2017dense} and YouCookII~\cite{Zhou2017TowardsAL}. 
Both automatic evaluation and human evaluation show that MART generates more satisfying results than previous LSTM-based approaches~\cite{Xiong2018MoveFA, Zhou2018GroundedVD, zhang2018cross} and transformer-based approaches~\cite{zhou2018end, dai2019transformer}.
In particular, MART can generate more coherent (e.g., coreference and order), less redundant paragraphs without losing paragraph accuracy (visual relevance).

\section{Related Work}\label{related_work}
\paragraph{Video Captioning}
Recently, video captioning has attracted much attention from both the computer vision and the natural language processing community.
Methods for the task share the same intrinsic nature of taking a video as the input and outputting a language description that can best describe the content, though they differ from each other on whether a single sentence~\cite{Wang_2019_ICCV, xu2016msr, chen2011collecting, pasunuru2017multi} or multiple sentences~\cite{rohrbach2014coherent, krishna2017dense, Xiong2018MoveFA, zhou2018end, gella2018dataset, park2019adversarial} are generated for the given video. 
In this paper, our goal falls into the category of generating a paragraph (multiple sentences) conditioned on an input video with several pre-defined event segments.

One line of work~\cite{zhou2018end, Zhou2018GroundedVD} addresses the video paragraph captioning task by decoding each video event segment separately into a sentence. 
The final paragraph description is obtained by concatenating the generated single sentence descriptions. Though individual sentences may precisely describe the corresponding event segments, when put together the sentences often become inconsistent and redundant.
Another line of works~\cite{Xiong2018MoveFA, gella2018dataset} use the LSTM decoder's last (word) hidden state from the previous sentence as the initial hidden state for the next sentence decoding, thus enabling information flow from previous sentences to subsequent sentences.
While these methods have shown better performance than their single sentence counterpart, they are still undesirable as the sentence-level recurrence is achieved at word-level, and the context history information quickly decays due to vanishing gradients~\cite{pascanu2013difficulty} problem.
Additionally, these designs also have difficulty modeling long-term dependencies~\cite{hochreiter2001gradient}. 
In comparison, the recurrence in MART resides in the sentence or segment level and is thus more robust to the aforementioned problems. 
AdvInf~\cite{park2019adversarial} augments the above LSTM word-level recurrence methods with adversarial inference, using a set of separately trained discriminators to re-rank the generated sentences. The techniques in AdvInf can be viewed as an orthogonal way of generating captions with better quality.

\paragraph{Transformers}
Transformer~\cite{vaswani2017attention} is used as the basis of our approach. 
Different from RNNs (e.g., LSTM~\cite{hochreiter1997long}, GRU~\cite{chung2014empirical}, etc) that use recurrent structure to model long-term dependencies, transformer relies on self-attention to learn the dependencies between input words. 
Transformers have proven to be more efficient and powerful than RNNs, with superior performance in many sequential modeling tasks, including machine translation~\cite{vaswani2017attention}, language modeling/pre-training~\cite{devlin2018bert, dai2019transformer, yang2019xlnet} and multi-modal representation learning~\cite{tan2019lxmert, chen2019uniter, sun2019videobert}. Additionally, \citet{zhou2018end} have shown that a transformer model can generate better captions than the LSTM model. 

However, transformer architectures are still unable to model history information well.  
This problem is identified in the task of language modeling as context fragmentation~\cite{dai2019transformer}, i.e., each language segment is modeled individually without knowing its surrounding context, leading to inefficient optimization and inferior performance.
To resolve this issue, Transformer-XL~\cite{dai2019transformer} introduces the idea of recurrence to the transformer language model.
Specifically, the modeling of a new language segment in Transformer-XL is conditioned on hidden states from previous language segments. 
Experimental results show Transformer-XL has stronger language modeling capability than the non-recurrent transformer.
Transformer-XL directly uses all the hidden states from the previous segment to enable recurrence. 
In comparison, our MART uses highly summarized memory states, making it more efficient in passing useful semantic or linguistic cues to future sentences. 

\section{Methods}\label{sec:methods}
Though our method provides a general temporal multi-modal learning framework, we focus on the video paragraph captioning task in this paper. 
Given a video $V$, with several temporally ordered event segments $[e_1, e_2, ..., e_T]$, the task is to generate a coherent paragraph consisting of multiple sentences $[s_1, s_2, ..., s_T]$ to describe the whole video, where sentence $s_t$ should describe the content in the segment $e_t$. 
In the following, we first describe the baseline transformer that generates sentences without recurrent architecture, then introduce our approach -- Memory-Augmented Recurrent Transformer (MART). 
Besides, we also compare MART with the recently proposed Transformer-XL~\cite{dai2019transformer} in detail.

\begin{figure}[t]
  \centering
  \includegraphics[width=0.9\linewidth]{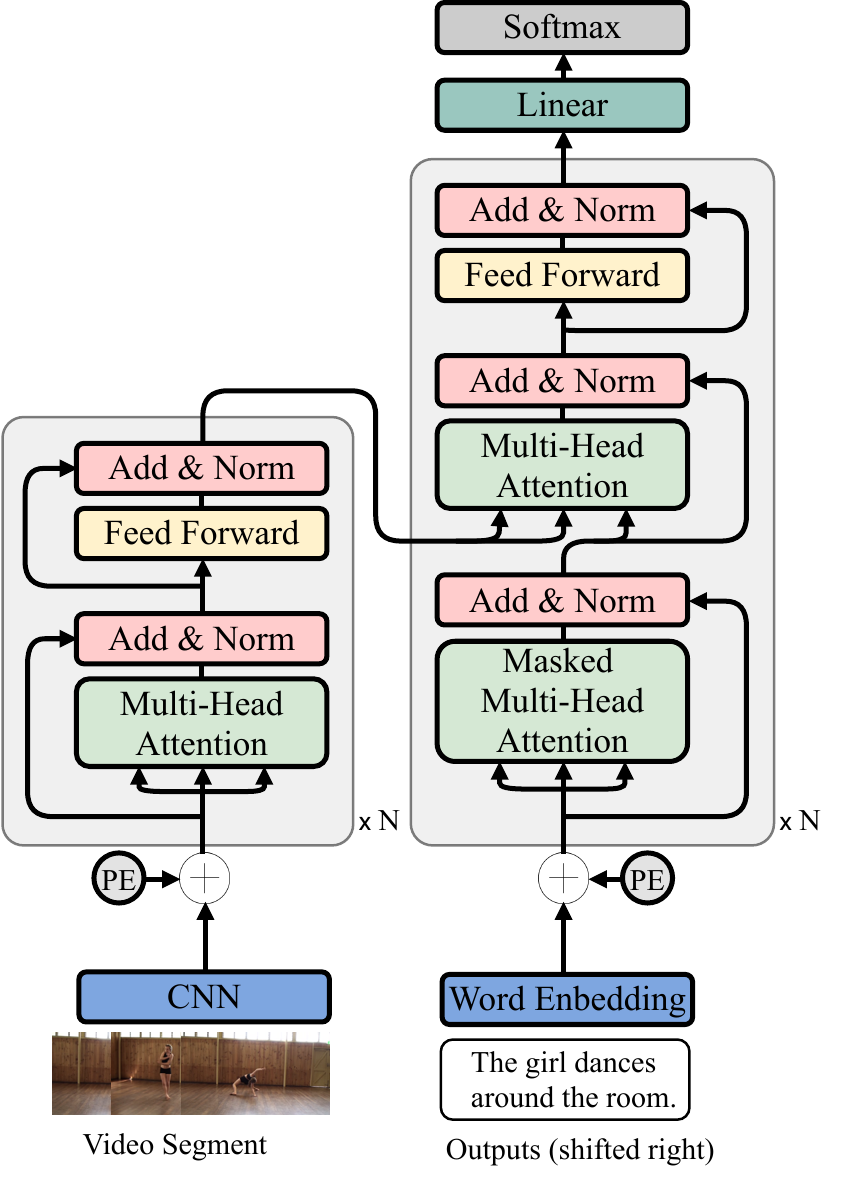}
  \caption{Vanilla transformer video captioning model~\cite{zhou2018end}. \textit{PE} denotes Positional Encoding, \textit{TE} denotes token Type Embedding.
  }
  \label{fig:vanilla_transformer}
\end{figure}

\begin{figure*}[!ht]
  \centering
  \includegraphics[width=\linewidth]{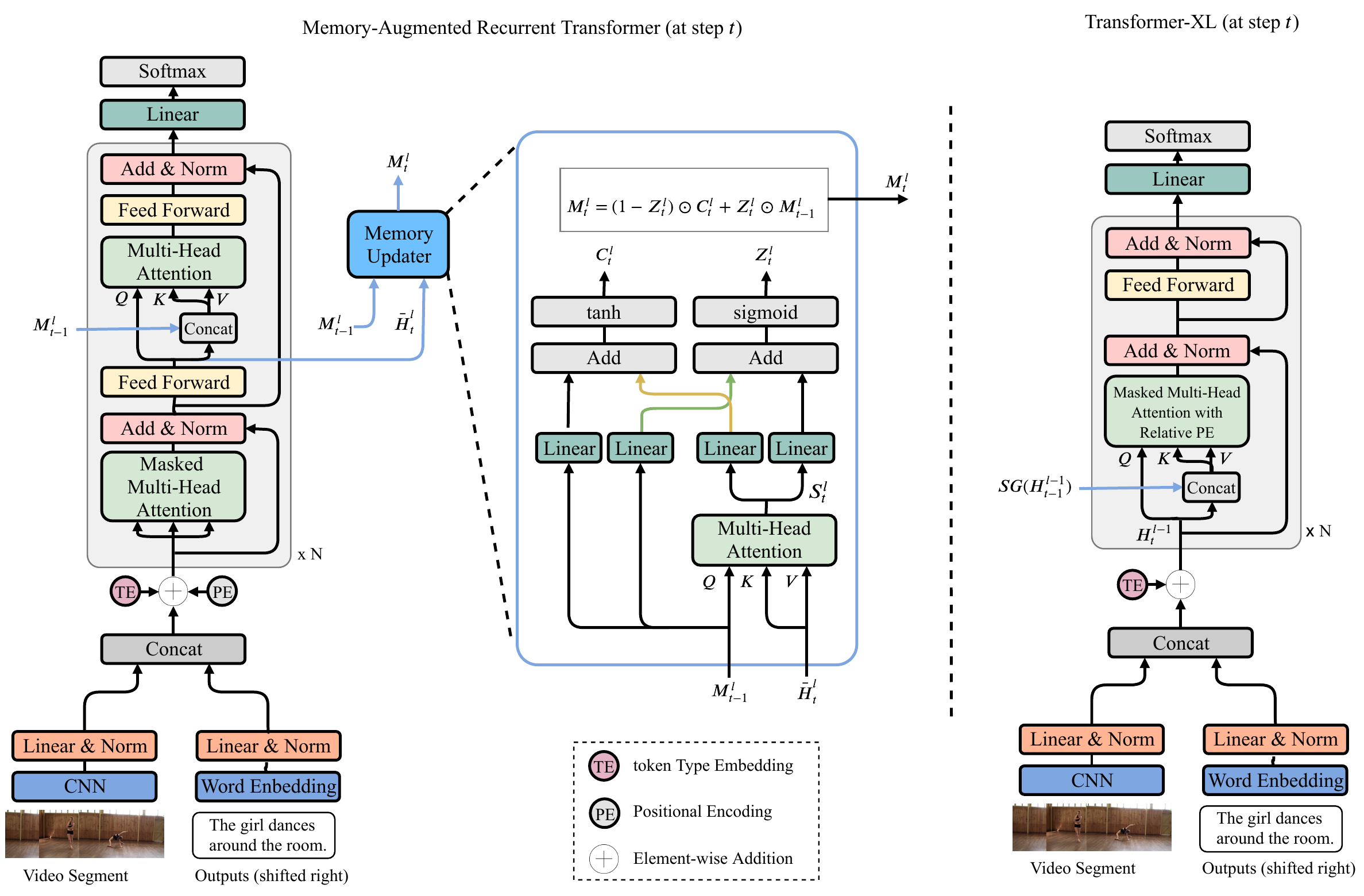}
  \caption{\textit{Left}: Our proposed Memory-Augmented Recurrent Transformer (MART) for video paragraph captioning. \textit{Right}: Transformer-XL~\cite{dai2019transformer} model for video paragraph captioning. \textit{Relative PE} denotes Relative Positional Encoding~\cite{dai2019transformer}.
  \textit{SG}($\cdot$) denotes stop-gradient, $\odot$ denotes Hadamard product. 
  }
  \label{fig:recurrent_transformers}
\end{figure*}

\subsection{Background: Vanilla Transformer}\label{subsec:vanilla_transformer}
We start by introducing the vanilla transformer video paragraph captioning model proposed by~\citet{zhou2018end}, which is an application of the original transformer~\cite{vaswani2017attention} model for video paragraph captioning. 
An overview of the model is shown in Figure~\ref{fig:vanilla_transformer}. 
The core of the architecture is the \textit{scaled dot-product attention}. 
Given query matrix $Q \in \mathbb{R}^{T_q \times d_k}$, key matrix $K \in \mathbb{R}^{T_v \times d_k}$ and value matrix $V \in \mathbb{R}^{T_v \times d_v}$, the attentional output is computed as:
\begin{align*}
    A(Q, K, V) = \textrm{softmax} \left( \frac{Q K^\top}{\sqrt{d_k}} , \textrm{dim}\mbox{=}1 \right) V,
\end{align*}
where $\textrm{softmax}(\cdot, \textrm{dim}\mbox{=}1)$ denotes performing softmax at the second dimension of the the input.
Combining \textit{h} paralleled scaled dot-product attention, we obtain the \textit{multi-head attention}~\cite{vaswani2017attention}, we denote it as \textit{MultiHeadAtt(Q, K, V)}.
The attention formulation discussed above is quite general. It can be used for various purposes, such as self-attention~\cite{vaswani2017attention} where query, key, and value matrix are all the same, and cross-attention~\cite{vaswani2017attention} where the query matrix is different from the key and value matrix. 
In this paper, we also use multi-head attention for memory aggregation and update, as discussed later.

The vanilla transformer video paragraph captioning model has $N$ encoder layers and $N$ decoder layers. At the $l$-th encoder layer, the multi-head attention module takes the last layer's hidden states $H^{l-1}$ as inputs and performs self-attention. The attentional outputs are then projected by a feed-forward layer. 
At the $l$-th decoder layer, the model first encodes the last decoder layer's hidden states using \textit{masked multi-head attention}.\footnote{\textit{masked multi-head attention} is used to prevent the model from seeing future words~\cite{vaswani2017attention}.}
It then uses multi-head attention, with the masked outputs as query matrix, and the hidden states $H^l$ from $l$-th encoder layer as key and value matrix to gather information from the encoder side. 
Similarly, a feed-forward layer is used to encode the sentences further. 
Residual connection~\cite{he2016deep} and layer-normalization~\cite{ba2016layer} are applied for each layer, for both encoder and decoder.

\subsection{Memory-Augmented Recurrent Transformer}
The vanilla transformer captioning model follows the classical encoder-decoder architecture, where the encoder and decoder network are separated. 
In comparison, the encoder and decoder are shared in MART, as shown in Figure~\ref{fig:recurrent_transformers} (\textit{left}).
The video and text inputs are firstly separately encoded and normalized. We denote the encoded video and text embeddings as $H^{0}_{video} \in  \mathbb{R}^{T_{video} \times d}$ and $H^{0}_{text} \in  \mathbb{R}^{T_{text} \times d}$, where $T_{video}$ and $T_{text}$ are the lengths of video and text, respectively. $d$ denotes the hidden size. 
We then concatenate these two embeddings as input to the transformer layers: $H^{0} \mbox{=} [H^{0}_{video}; H^{0}_{text}] \in \mathbb{R}^{T_{c} \times d}$, where $[;]$ denotes concatenation, $T_{c} \mbox{=} T_{video} + T_{text}$.
This unified encoder-decoder design is inspired by recent works on multi-modal representation learning~\cite{chen2019uniter, sun2019videobert}.
We also use two trainable token type embedding vectors to indicate whether an input token is from video or text, similar to~\citet{devlin2018bert} where the token type embeddings are added to indicate different input sequences. 
We ignore the video token positions and only consider the text token positions when calculating loss and generating words.

While the aforementioned vanilla transformer is a powerful method, it is less suitable for video paragraph captioning due to its inability to utilize video segments and sentences history information. 
Thus, given the unified encoder-decoder transformer, we augment it with an external memory module, which helps it to utilize video segments and the corresponding caption history to generate the next sentence.
An overview of the memory module is shown in Figure~\ref{fig:recurrent_transformers} (\textit{left}).
At step $t$, i.e., decoding the $t$-th video segment, the $l$-th layer aggregates the information from both its intermediate hidden states $\bar{H}_{t}^{l} \in \mathbb{R}^{T_{c} \times d}$ and the memory states $M_{t-1}^{l} \in \mathbb{R}^{T_m \times d}$ ($T_m$ denotes memory state length or equivalently \#slots in the memory) from the last step, using a multi-head attention. 
The input query matrix of the multi-head attention $Q \mbox{=} \bar{H}_{t}^{l}$, key and value matrices are $K, V \mbox{=} [M_{t-1}^{l}; \bar{H}_{t}^{l}] \in \mathbb{R}^{(T_m + T_c) \times d}$. 
The memory augmented hidden states are further encoded using a feed forward layer and then merged with the intermediate hidden states $\bar{H}_{t}^{l}$ using a residual connection and layer norm to form the hidden states output $H_{t}^{l} \in \mathbb{R}^{T_{c} \times d}$.
The memory state $M_{t-1}^l$ is updated as $M_{t}^l$, using the intermediate hidden states $\bar{H}_{t}^{l}$. This process is conducted in the \textit{Memory Updater} module, illustrated in Figure~\ref{fig:recurrent_transformers}. We summarize the procedure below:
\vspace{3pt}
\begin{align*}
    S_{t}^{l} &= \textrm{MultiHeadAtt}(M_{t-1}^{l}, \bar{H}_{t}^{l}, \bar{H}_{t}^{l}), \\
    C_{t}^{l} &= \textrm{tanh}(W_{mc}^{l} M_{t-1}^{l} + W_{sc}^{l} S_{t}^{l} + b_{c}^{l}), \\
    Z_{t}^{l} &= \textrm{sigmoid}(W_{mz}^{l} M_{t-1}^{l} + W_{sz}^{l} S_{t}^{l} + b_{z}^{l}), \\
    M_{t}^{l} &= (1 - Z_{t}^{l}) \odot C_{t}^{l} + Z_{t}^{l} \odot M_{t-1}^{l},
\end{align*}
\vspace{3pt}
\noindent where $\odot$ denotes Hadamard product, $W_{mc}^{l}$, $W_{sc}^{l}$, $W_{mz}^{l}$, and $ W_{sz}^{l}$ are trainable weights, $b_{c}^{l}$ and $b_{z}^{l}$ are trainable bias. $C_{t}^{l} \in \mathbb{R}^{T_m \times d}$ is the internal cell state. 
$Z_{t}^{l} \in \mathbb{R}^{T_m \times d}$ is the update gate that controls which information to retain from the previous memory state, and thus reducing redundancy and maintaining coherence in the generated paragraphs. 

This update strategy is conceptually similar to LSTM~\cite{hochreiter1997long} and GRU~\cite{chung2014empirical}. 
It differs in that multi-head attention is used to encode the memory state and thus multiple memory slots are supported instead of a single one in LSTM and GRU, which gives it a higher capacity of modeling complex relations.
Recent works~\cite{sukhbaatar2015end,graves2014neural,xiong2016dynamic} introduce a memory component into neural networks, where the memory is mainly designed to memorize facts in the input context to support downstream tasks, e.g., copy~\cite{graves2014neural} or question answering~\cite{sukhbaatar2015end,xiong2016dynamic}. In comparison, the memory in MART is designed to memorize the sequence generation history to support the coherent generation of the next sequence.

\subsection{Comparison with Transformer-XL}\label{subsec:transformer-xl}
\textit{Transformer-XL}~\cite{dai2019transformer} is a recently proposed transformer-based language model that uses a segment-level recurrence mechanism to capture the long-term dependency in context. 
In Figure~\ref{fig:recurrent_transformers} (\textit{right}) we show a modified version of Transformer-XL for video paragraph captioning. 
At step $t$, at its $l$-th layer, Transformer-XL takes as inputs the last layer's hidden states from both the current step and the last step, which we denote as $H_{t}^{l-1}$ and $SG(H_{t-1}^{l-1})$, where $SG(\cdot)$ stands for stop-gradient, and is used to save GPU memory and computation~\cite{dai2019transformer}.  
The input query matrix of the multi-head attention $Q = H_{t}^{l-1}$, key and value matrices are $K, V = [SG(H_{t-1}^{l-1}); H_{t}^{l-1}]$.  
Note the multi-head attention here is integrated with relative positional encoding~\cite{dai2019transformer}.

Both designed to leverage the long-term dependency in context, the recurrence in Transformer-XL is between $H_{t}^{l}$ and $H_{t-1}^{l-1}$, which shifts one layer downwards per step. 
This mismatch in representation granularity may potentially be harmful to the learning process and affect the model performance.
In contrast, the recurrence in MART is between $\bar{H}_{t}^{l}$ and $M_{t-1}^{l}$ (updated using $\bar{H}_{t-1}^{l}$) of the same layer.
Besides, Transformer-XL directly uses all the hidden states from the last step to enable recurrence, which might be less effective as less relevant and repetitive information is also passed along. 
In comparison, MART achieves recurrence by using memory states that are highly summarized from previous steps, which may help the model to reduce redundancy and only keep important information from previous steps.

\section{Experiments}\label{experiments}
We conducted experiments on two popular benchmark datasets, ActivityNet Captions~\cite{krishna2017dense} and YouCookII~\cite{Zhou2017TowardsAL}. 
We evaluate our proposed MART and compare it with various baseline approaches.

\subsection{Data and Evaluation Metrics} \label{subsec:dataset_and_metric}
\paragraph{Datasets} ActivityNet Captions~\cite{krishna2017dense} contains 10,009 videos in \textit{train} set, 4,917 videos in \textit{val} set. 
Each video in \textit{train} has a single reference paragraph while each video in \textit{val} has two reference paragraphs.
\citet{park2019adversarial} uses the same set of videos (though different segments) in \textit{val} for both validation and test.
To allow better evaluation of the models, we use splits provided by~\citet{Zhou2018GroundedVD}, where the original \textit{val} set is split into two subsets: \textit{ae-val} with 2,460 videos for validation and \textit{ae-test} with 2,457 videos for test.
This setup makes sure the videos used for test will not be seen in validation.
YouCookII~\cite{Zhou2017TowardsAL} contains 1,333 training videos and 457 validation videos. Each video has a single reference paragraph.
Both datasets come with temporal event segments annotated with human written natural language sentences. 
On average, there are 3.65 event segments for each video in ActivityNet Captions, 7.7 segments for each video in YouCookII.

\paragraph{Data Preprocessing} We use aligned appearance and optical flow features extracted at 2FPS to represent videos, provided by~\citet{zhou2018end}.
Specifically, for appearance, 2048D feature vectors from the `Flatten-673' layer in ResNet-200~\cite{he2016deep} are used; 
for optical flow, 1024D feature vectors from the `global pool' layer of  BN-Inception~\cite{ioffe2015batch} are used. 
Both networks are pre-trained on ActivityNet~\cite{caba2015activitynet} for action recognition, provided by~\cite{xiong2016cuhk}. 
We truncate sequences longer than 100 for video and 20 for text and set the maximum number of video segments to 6 for ActivityNet Captions and 12 for YouCookII. 
Finally, we build vocabularies based on words that occur at least 5 times for ActivityNet Captions and 3 times for YouCookII. The resulting vocabulary contains 3,544 words for ActivityNet Captions and 992 words for YouCookII.

\paragraph{Evaluation Metrics (Automatic and Human)} 
We evaluate the captioning performance at paragraph-level, following~\cite{park2019adversarial, Xiong2018MoveFA}, reporting numbers on standard metrics, including BLEU@4~\cite{papineni2002bleu},  METEOR~\cite{denkowski2014meteor}, CIDEr-D~\cite{vedantam2015cider}. 
Since these metrics mainly focus on whether the generated paragraph matches the ground-truth paragraph, they fail to evaluate the redundancy of these multi-sentence paragraphs. 
Thus, we follow previous works~\cite{park2019adversarial, Xiong2018MoveFA} to evaluate repetition using R@4. It measures the degree of N-gram (N=4) repetition in the descriptions. 
Besides the automated metrics, we also conduct human evaluations to provide additional comparisons between the methods. We consider two aspects in human evaluation, \textit{relevance} (i.e., how related is a generated paragraph caption to the content of the given video) and \textit{coherence} (i.e., whether a generated paragraph caption reads fluently and is linguistically coherent over its multiple sentences).

\begin{table*}[!t]
\centering
\small
\scalebox{1.0}{
\begin{tabular}{lccccccccc}
\toprule
\multirow{2}{*}{Model} & \multirow{2}{*}{Re.} & \multicolumn{4}{c}{ActivityNet Captions (\textit{ae-test})} & \multicolumn{4}{c}{YouCookII (\textit{val})} \\  
\cmidrule(rl){3-6}  \cmidrule(rl){7-10}
& & B@4 & M & C & R@4~$\downarrow$ & B@4 & M & C & R@4~$\downarrow$ \\ 
\midrule
VTransformer~\shortcite{zhou2018end}  & \xmark & 9.31 & 15.54 & 21.33 & 7.45 & 7.62 & 15.65 & 32.26 & 7.83 \\
Transformer-XL~\shortcite{dai2019transformer}  & \cmark & \textbf{10.25} & 14.91 & 21.71 & 8.79 & 6.56 & 14.76 & 26.35 & 6.30 \\
Transformer-XLRG   &\cmark & 10.07 & 14.58 & 20.34 & 9.37 & 6.63 & 14.74 & 25.93 & 6.03 \\
MART  &\cmark & 9.78  & \textbf{15.57} & \textbf{22.16} & \textbf{5.44}   & \textbf{8.00} & \textbf{15.9} & \textbf{35.74} & \textbf{4.39}  \\
\midrule
Human & - & - & - & -  & 0.98 & - & - & - & 1.27 \\
\bottomrule
\end{tabular}
}
\caption{Comparison with transformer baselines on ActivityNet Captions \textit{ae-test} split and YouCookII \textit{val} split. Re. indicates whether sentence-level recurrence is used. We report BLEU@4 (B@4), METEOR (M), CIDEr-D (C) and Repetition (R@4). \textit{VTransformer} denotes vanilla transformer.
} 
\label{tab:main_res}
\end{table*}

\begin{table}[!ht]
\centering
\small
\setlength{\tabcolsep}{4pt}
\scalebox{0.84}{
\begin{tabular}{lcccccc}
\toprule
& Det. & Re. &  B@4 & M & C & R@4~$\downarrow$ \\ \midrule
\multicolumn{7}{l}{\textbf{LSTM based methods}}\\
MFT~\shortcite{Xiong2018MoveFA} & \xmark & \cmark & 10.29 & 14.73 & 19.12 & 17.71 \\
HSE~\shortcite{zhang2018cross} & \xmark & \cmark & 9.84 & 13.78 & 18.78 & 13.22 \\
\midrule
\multicolumn{7}{l}{\textbf{LSTM based methods with detection feature}} \\
\rowcolor{Gray}
GVD~\shortcite{Zhou2018GroundedVD} & \cmark & \xmark & 11.04 & 15.71 & 21.95 & 8.76 \\
\rowcolor{Gray}
\rowcolor{Gray}
GVDsup~\shortcite{Zhou2018GroundedVD} & \cmark & \xmark  & \textbf{11.30} & 16.41 & 22.94  & 7.04 \\
\rowcolor{Gray}
AdvInf~\shortcite{park2019adversarial} & \cmark & \cmark & 10.04 & \textbf{16.60} & 20.97 & 5.76 \\
\midrule
\multicolumn{7}{l}{\textbf{Transformer based methods}} \\
VTransformer~\shortcite{zhou2018end} & \xmark  & \xmark & 9.75 & 15.64 & 22.16 & 7.79 \\
Transformer-XL~\shortcite{dai2019transformer} & \xmark & \cmark & 10.39 & 15.09 & 21.67 & 8.54  \\
Transformer-XLRG & \xmark  & \cmark & 10.17 & 14.77 & 20.40 & 8.85   \\
(Ours) MART & \xmark  & \cmark & 10.33  & 15.68 & \textbf{23.42} & \textbf{5.18} \\
\midrule
Human & - & - & - & - & - & 0.98 \\
\bottomrule
\end{tabular}
}
\caption{Comparison with baselines on ActivityNet Captions \textit{ae-val} split. \textit{Det.} indicates whether the model uses detection feature. Models that use detection features are shown in gray background to indicate they are not in fair comparison with the others. \textit{Re.} indicates whether sentence-level recurrence is used. \textit{VTransformer} denotes vanilla transformer. 
}
\label{tab:anet_val_res}
\end{table}

\subsection{Implementation Details}
MART is implemented in PyTorch~\cite{paszke2017automatic}. We set the hidden size to 768, the number of transformer layers to 2, and the number of attention heads to 12. For positional encoding, we follow~\citet{vaswani2017attention} to use the fixed scheme. For memory module, we set the length of recurrent memory state to 1, i.e., $T_m\mbox{=}1$.
We optimize the model following the strategy used by~\citet{devlin2018bert}. 
Specifically, we use Adam~\cite{kingma2014adam} with an initial learning rate of 1e-4, $\beta_1\mbox{=}0.9$, $\beta_2\mbox{=}0.999$, L2 weight decay of 0.01, and learning rate warmup over the first 5 epochs.
We train the model for at most 50 epochs with early stopping using CIDEr-D and batch size 16. We use greedy decoding as we did not observe better performance using beam search.

\subsection{Baselines}
\paragraph{Vanilla Transformer} 
This model originates from the transformer~\cite{vaswani2017attention}, proposed by~\citet{zhou2018end} (more details in Section~\ref{subsec:vanilla_transformer}). 
It takes a single video segment as input and independently generates a single sentence describing the given segment. 
Note that \citet{zhou2018end} also have a separate proposal generation module, but here we only focus on its captioning module. 
To obtain paragraph-level captions, the independently generated single sentence captions are concatenated as the output paragraph.

\paragraph{Transformer-XL} 
Transformer-XL is proposed by~\citet{dai2019transformer} for modeling long-term dependency in natural language. Here we adapt it for video paragraph captioning (more details in Section~\ref{subsec:transformer-xl}). 
The original design of Transformer-XL stops gradients from passing between different recurrent steps to save GPU memory and computation. To enable a more fair comparison with our model, we implemented a version that allows gradient flow through different steps, calling this \textit{Transformer-XLRG} (Transformer-XL with Recurrent Gradient). 

\paragraph{AdvInf}
AdvInf~\cite{park2019adversarial} uses a set of three discriminators to do adversarial inference on a strong LSTM captioning model. 
The input features of the LSTM model are the concatenation of image recognition, action recognition, and object detection features. 
To encourage temporal coherence between consecutive sentences, the last hidden state from the previous sentence is used as input to the decoder~\cite{Xiong2018MoveFA, gella2018dataset}. 
The three discriminators are trained adversarially and are specifically designed to reduce repetition and encourage fluency and relevance in the generated paragraph.

\paragraph{GVD} 
An LSTM based model for grounded video description~\cite{Zhou2018GroundedVD}. 
It uses densely detected object regions as inputs, with a grounding module that grounds generated words to the regions. 
Additionally, we also consider a GVD variant (\textit{GVDsup}) that uses grounding supervision from~\citet{Zhou2018GroundedVD}. 

\paragraph{MFT} 
MFT~\cite{Xiong2018MoveFA} uses an LSTM model with a similar sentence-level recurrence as in AdvInf~\cite{park2019adversarial}. 

\paragraph{HSE} 
HSE~\cite{zhang2018cross} is a hierarchical model designed to learn both clip-sentence and paragraph-video correspondences. Given the learned contextualized video embedding, HSE uses a 2-layer LSTM to generate captions.

For AdvInf, MFT, HSE, GVD, and GVDsup, we obtain generated sentences from the authors. 
We only report their performance on ActivityNet Captions \textit{ae-val} split to enable a fair comparison, as $(i)$ AdvInf, MFT and HSE have different settings as ours, where \textit{ae-test} videos are included as part of their validation set; $(ii)$ we do not have access to the \textit{ae-test} predictions of GVD and GVDsup.
For vanilla transformer, Transformer-XL and Transformer-XLRG, we borrow/modify the model implementations from the original authors and train them under the same settings as MART.

\subsection{Results}\label{subsec:results}
\paragraph{Automatic Evaluation}
Table~\ref{tab:main_res} shows the results of MART and several transformer baseline methods. 
We observe stronger or comparable performance for the language metrics (B@4, M, C) for both ActivityNet Captions and YouCookII datasets.
For R@4, MART produces significantly better results compared to the three transformer baselines, showing its effectiveness in reducing redundancy in the generated paragraphs.
Table~\ref{tab:anet_val_res} shows the comparison of MART with state-of-the-art models on ActivityNet Captions. 
MART achieves the best scores for both CIDEr-D and R@4 and has a comparable performance for B@4 and METEOR. 
Note that the best B@4 model, GVDsup~\cite{Zhou2018GroundedVD}, and the best METEOR model, AdvInf~\cite{park2019adversarial}, both use strong detection features, and GVDsup has also used grounding supervision. 
Regarding the repetition score R@4, MART has the highest score. It outperforms the strong adversarial model AvdInf~\cite{park2019adversarial} even in an unfair comparison where AdvInf uses extra detection features. 
Additionally, AdvInf has a time-consuming adversarial training and decoding process where a set of discriminator models are trained and used to re-rank candidate sentences, while MART can do much faster inference with only greedy decoding and no further post-processing.
The comparisons in Table~\ref{tab:main_res} and Table~\ref{tab:anet_val_res} show that MART is able to generate less redundant (thus more coherent) paragraphs while maintaining relevance to the videos.

\begin{figure*}[!t]
  \centering
  \includegraphics[width=1\linewidth]{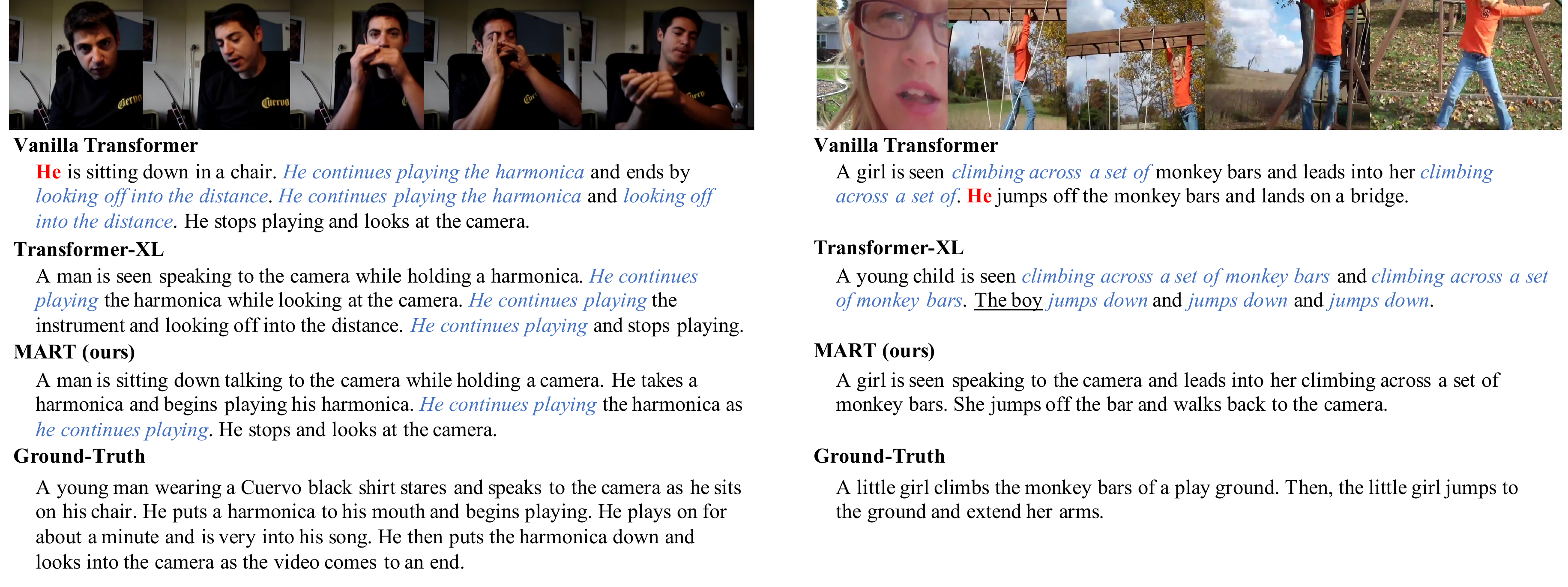}
  \caption{Qualitative examples. Red/bold indicates pronoun errors (inappropriate use of pronouns), \text{blue/italic} indicates repetitive patterns, underline indicates content errors. Compared to baselines, our model generates more coherent, less repeated paragraphs while maintaining relevance. 
  }
  \label{fig:caption_example}
\end{figure*}

\begin{table}[!t]
\centering
\small
\scalebox{0.82}{
\begin{tabular}{lccc}
\toprule
 \makecell{} & \makecell{MART wins (\%)} & \makecell{VTransformer wins (\%)} &  \makecell{Delta} \\ \midrule
relevance & 37 & 29.5 & \textbf{+7.5} \\
coherence & 42.8 & 26.3 & \textbf{+16.5}\\
\bottomrule 
& & & \\
\toprule
 \makecell{} & \makecell{MART wins (\%)} & \makecell{Transformer-XL wins (\%)} &  \makecell{Delta}  \\ \midrule
relevance & 40.0 & 39.5 & \textbf{+0.5} \\
coherence & 39.2 & 36.2 & \textbf{+3.0} \\
\bottomrule
\end{tabular}
}
\caption{Human evaluation on ActivityNet Captions \textit{ae-test} set w.r.t. relevance and coherence. \textit{Top}: MART \textit{vs.} vanilla transformer (VTransformer). \textit{Bottom}: MART \textit{vs.} Transformer-XL.}
\label{tab:huamn_eval}
\end{table}

\paragraph{Human Evaluation}
In addition to the automatic metrics, we also run human evaluation on Amazon Mechanical Turk (AMT) with 200 randomly sampled videos from ActivityNet Captions \textit{ae-test} split, where each video was judged by three different AMT workers.
We design a set of pairwise experiments~\cite{pasunuru2017reinforced, park2019adversarial}, where we compare two models at a time. 
AMT workers are instructed to choose which caption is better or the two captions are not distinguishable based on relevance and coherence, respectively. 
The models are anonymized, and the predictions are shuffled. In total, we have 54 workers participated the MART \textit{vs.} vanilla transformer experiments, 47 workers participated the MART \textit{vs.} Transformer-XL experiments.
In Table~\ref{tab:huamn_eval} we show human evaluation results, where the scores are calculated as the percentage of workers that have voted a certain option. 
With its sentence-level recurrence mechanism, MART is substantially better than the vanilla transformer model for both relevance and coherence.
Compared to the strong baseline approach Transformer-XL, MART has similar performance in terms of relevance, but still reasonably better performance in terms of coherence.

\begin{table}[!t]
\centering
\small
\setlength{\tabcolsep}{4pt}
\scalebox{0.82}{
\begin{tabular}{lccccccc}
\toprule
& \makecell{\#hidden \\ layers} & \makecell{mem. \\ len.} & Re. &  B@4 & M & C & R@4~$\downarrow$ \\ 
\midrule
\multicolumn{3}{l}{\bf \#hidden layers} & & & &  \\
MART & 1  & 1 & \cmark & \bf 10.42  & \bf 16.01 & 22.87  & 6.70 \\
MART & 5  & 1 & \cmark & \bf 10.48  & \bf 16.03 & \bf 24.33  & 6.74 \\
\midrule
\multicolumn{3}{l}{\bf mem. len.} & & & &  \\
MART & 2  & 2 & \cmark & 10.30  & 15.66 & 22.93  & \bf 5.94 \\
MART & 2  & 5 & \cmark & 10.12 & 15.48 & 22.89  & 6.83 \\
\midrule
\multicolumn{3}{l}{\bf recurrence} & & & &  \\
MART w/o re. & 2  & - & \xmark & 9.91 & 15.83 & 22.78 & 7.56  \\
\midrule
MART & 2  & 1 & \cmark &   10.33  & 15.68 & \bf 23.42 & \bf 5.18 \\
\bottomrule
\end{tabular}
}
\caption{Model ablation on ActivityNet Captions \textit{ae-val} split. \textit{Re.} indicates whether sentence-level recurrence is used. \textit{mem. len.} indicates the length of the memory state. \textit{MART w/o re.} denotes a MART variant without recurrence. Top two scores are highlighted.}
\label{tab:anet_val_ablation}
\end{table}

\paragraph{Model Ablation}
We show model ablation in Table~\ref{tab:anet_val_ablation}. 
MART models with recurrence have better overall performance than the variant without, suggesting the effectiveness of our recurrent memory design.
We choose to use the model with 2 hidden layers and memory state length 1 as it shows a good balance between performance and computation.

\begin{figure}[!t]
  \centering
  \includegraphics[width=0.9\linewidth]{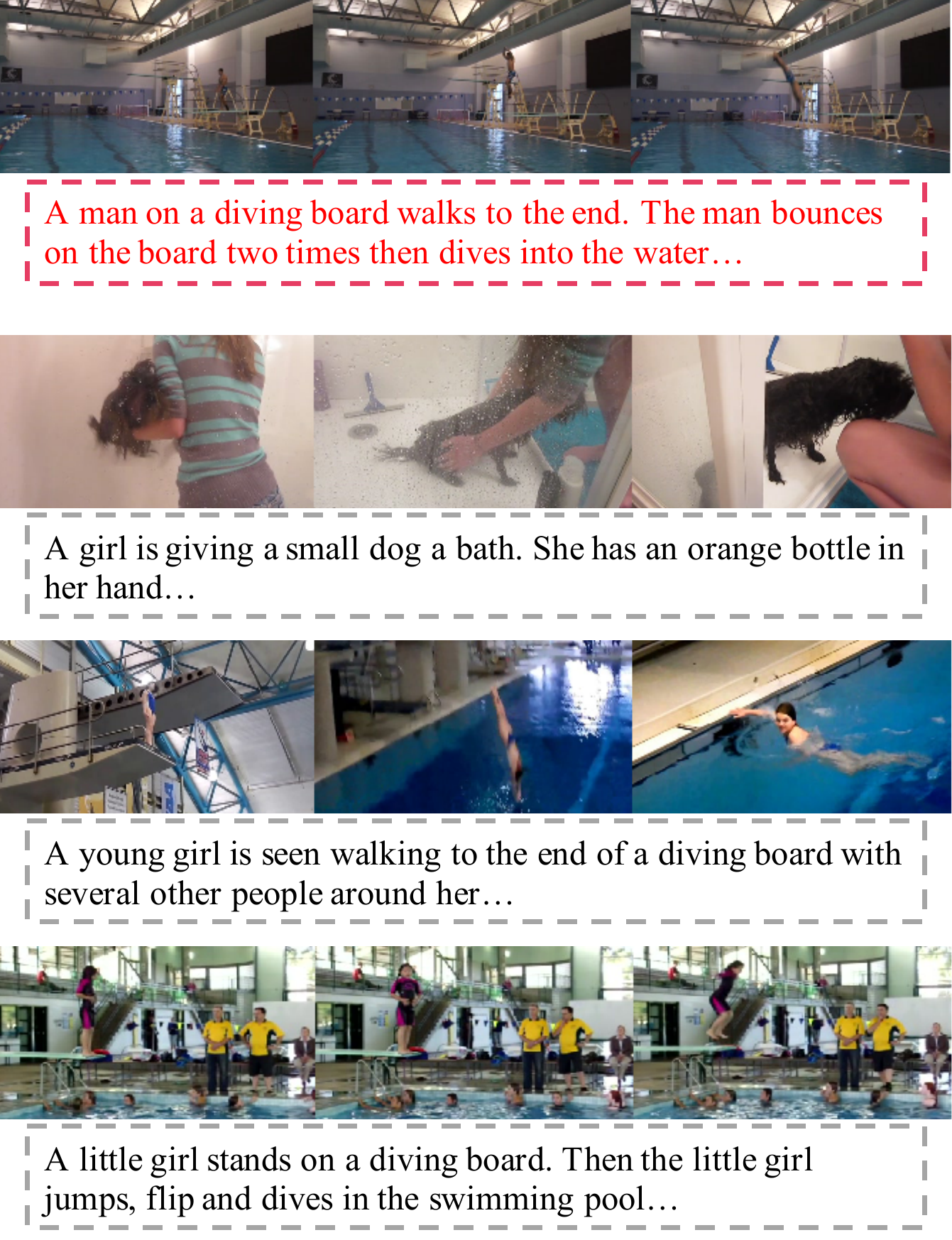}
  \caption{Nearest neighbors retrieved using memory states. Top row shows the query, the 3 rows below it are the top-3 nearest neighbors.}
  \label{fig:mem_retrieve_example}
\end{figure}

\paragraph{Qualitative Examples} 
In Figure~\ref{fig:caption_example}, we show paragraph captions generated by vanilla transformer, Transformer-XL, and our method MART. 
Compared to the two baselines, MART produces more coherent and less redundant paragraphs. 
In particular, we noticed that vanilla transformer often uses incoherent pronouns/person mentions, while MART and Transformer-XL is able to use suitable pronouns/person mentions across the sentences and thus improve the coherence of the paragraph.
Compare with Transformer-XL, we found that the paragraphs generated by MART have much less cross-sentence repetitions. 
We attribute MART's success to its recurrence design - the previous memory states are highly summarized, in which redundant information is removed.
While there is less redundancy between sentences generated by MART, in Figure~\ref{fig:caption_example} (\textit{left}), we noticed that repetition still exists within a single sentence, suggesting further efforts on reducing the repetition in single sentence generation. More examples are in the appendix.

\paragraph{Memory Ablation}
To explore whether the learned memory state could store useful information about the videos and captions, we conducted a video retrieval experiment on ActivityNet Captions \textit{train} split with 10K videos, where we extract the last step memory state in the first layer of a trained MART model for each video as its representation to perform nearest neighbor search with cosine similarity. 
Though not explicitly trained for the retrieval task, we observe some positive examples in the experiments. We show an example in Figure~\ref{fig:mem_retrieve_example}, the neighbors mostly show related activities.

\section{Conclusion}\label{conclusion}
In this work, we present a new approach -- Memory-Augmented Recurrent Transformer (MART) for video paragraph captioning, where we designed an auxiliary memory module to enable recurrence in transformers.
Experimental results on two standard datasets show that MART has better overall performance than the baseline methods. 
In particular, MART can generate more coherent, less redundant paragraphs without any degradation in relevance.

\section*{Acknowledgments}

We thank the anonymous reviewers for their helpful comments and discussions. This work was performed while Jie Lei was an intern at Tencent AI Lab, Seattle, USA. It was later partially supported by NSF Awards CAREER-1846185, 1562098, DARPA KAIROS Grant FA8750-19-2-1004, and ARO-YIP Award W911NF-18-1-0336. The views contained in this article are those of the authors and not of the funding agency.

\bibliography{acl2020}
\bibliographystyle{acl_natbib}

\appendix
\section{Appendices}\label{sec:appendix}

\subsection{Additional Qualitative Examples}
\begin{figure*}[t]
  \centering
  \includegraphics[width=\linewidth]{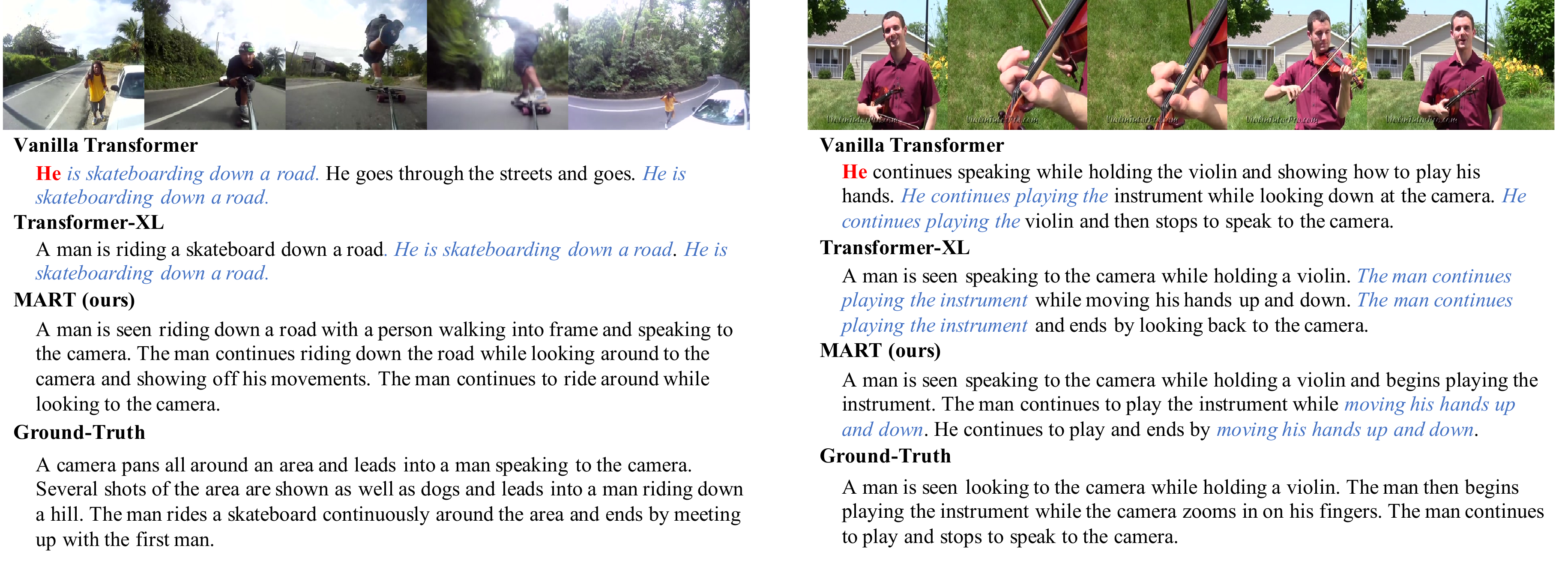} \\
  \includegraphics[width=\linewidth]{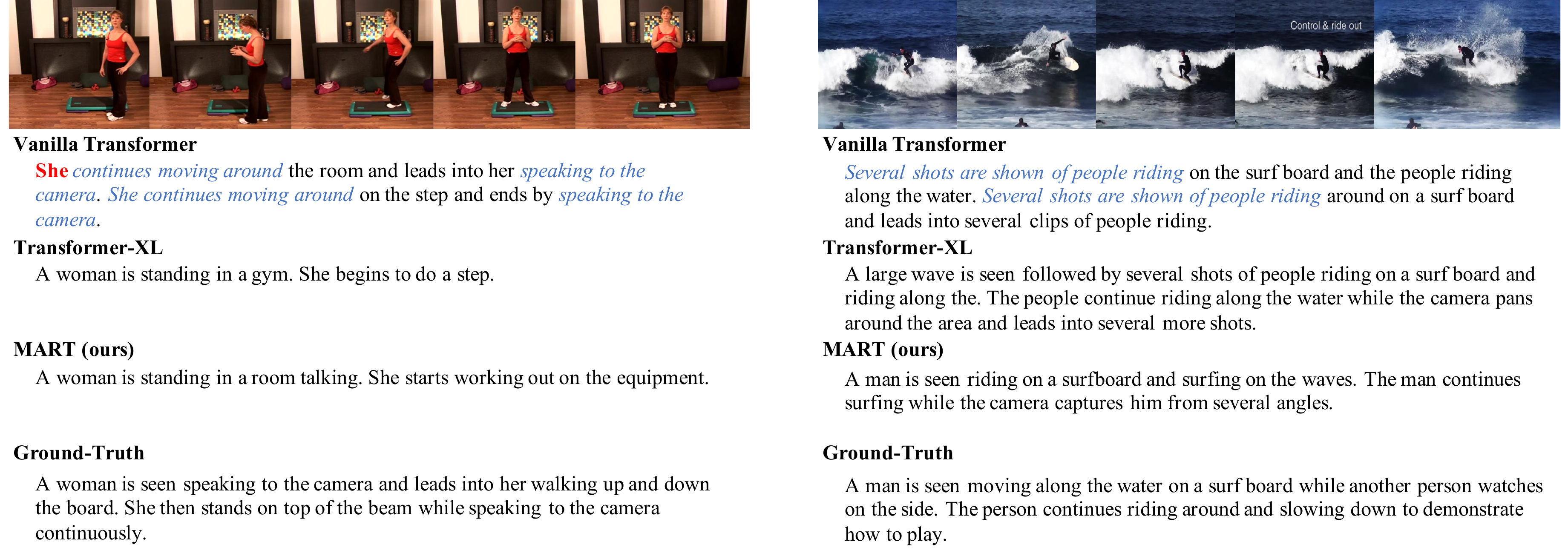} \\
  \includegraphics[width=\linewidth]{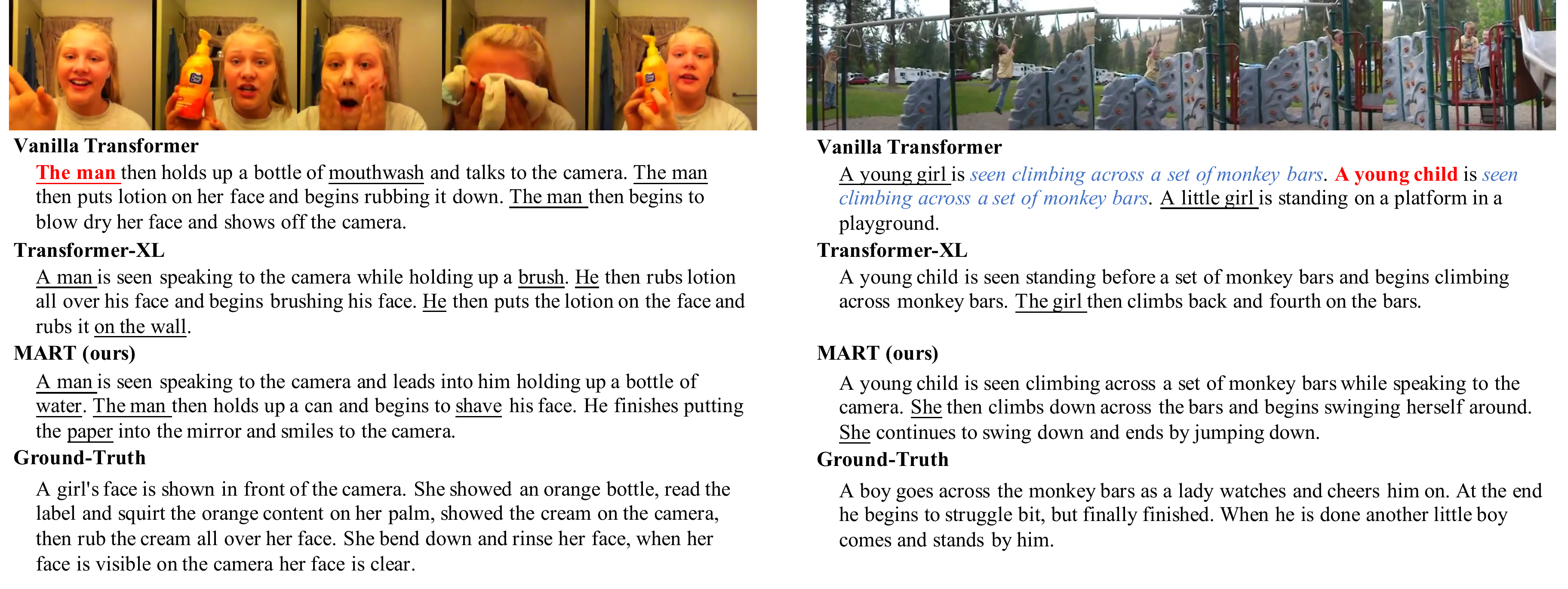}
  \caption{Additional qualitative examples. Red/bold indicates pronoun errors (inappropriate use of pronouns or person mentions), \text{blue/italic} indicates repetitive patterns, underline indicates content errors. Compared to baselines, our model generates more coherent, less repeated paragraphs while maintaining relevance.}
  \label{fig:more_caption_examples}
\end{figure*}

We show more caption examples in Figure~\ref{fig:more_caption_examples}.
Overall, we see captions generated by models with sentence-level recurrence, i.e., MART and Transformer-XL, tend to be more coherent. Comparing with Transformer-XL, captions generated by MART are usually less repetitive. However, as shown in the two examples at the last row of Figure~\ref{fig:more_caption_examples}, all three models suffer from the content error, where the models are not able to recognize and describe the fine-grained details in the videos, e.g., gender and fine-grained objects/actions.

\end{document}